\documentclass[10pt,twocolumn,letterpaper]{article}

\usepackage[pagenumbers]{cvpr} 

\usepackage{times}
\usepackage{epsfig}
\usepackage{graphicx}
\usepackage{amsmath}
\usepackage{amssymb}
\usepackage{subcaption}
\usepackage{multirow}
\usepackage{algorithm}
\usepackage[dvipsnames]{xcolor}
\usepackage[noend]{algpseudocode}
\usepackage[export]{adjustbox}
\usepackage{rotating}
\usepackage[accsupp]{axessibility}  

\usepackage[pagebackref,breaklinks,colorlinks]{hyperref}

\usepackage[capitalize]{cleveref}
\crefname{section}{Sec.}{Secs.}
\Crefname{section}{Section}{Sections}
\Crefname{table}{Table}{Tables}
\crefname{table}{Tab.}{Tabs.}

\def\cvprPaperID{7} 
\def\confName{CVPR}
\def\confYear{2023}

\begin{document}

\title{A Scale-Invariant Trajectory Simplification Method for Efficient Data Collection in Videos}

\author{Yang Liu\\
Magic Leap\\
{\tt\small yaliu@magicleap.com}
\and
Luiz G. Hafemann\\
Ubisoft La Forge\\
{\tt\small luiz@hafemann.ca} 
}


\twocolumn[{%
\renewcommand\twocolumn[1][]{#1}%
\maketitle

\vspace{-1.2cm}
\begin{center}
    \centering
    \captionsetup{type=figure}
    \includegraphics[width=\textwidth, trim={0in 0in 0in 0},clip]{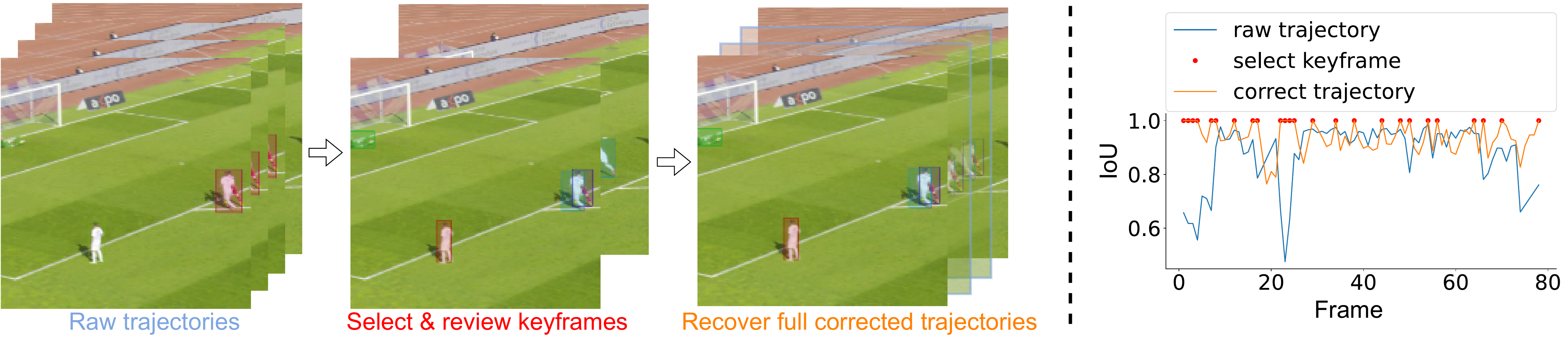}
    \vspace{-0.8cm}
    \captionof{figure}{Illustration of our method, where the left part showcases the correction framework and the right part depicts the IoU score between the ground truth trajectory and trajectory in each step of the framework  (represented by the corresponding color). More precisely, our approach can automatically choose crucial keyframes (\textcolor{red}{red}) for manual inspection and correction from the initial tracking trajectories (\textcolor{blue}{blue}), which helps to minimize the annotation expenses. After that, the precise trajectories (\textcolor{orange}{orange}) are reconstructed via interpolation.} 
    \label{fig:framework_eg}
\end{center}%
}]

\begin{abstract}
Training data is a critical requirement for machine learning tasks, and labeled training data can be expensive to acquire, often requiring manual or semi-automated data collection pipelines. For tracking applications, the data collection involves drawing bounding boxes around the classes of interest on each frame, and associate detections of the same ``instance" over frames. In a semi-automated data collection pipeline, this can be achieved by running a baseline detection and tracking algorithm, and relying on manual correction to add/remove/change bounding boxes on each frame, as well as resolving errors in the associations over frames (track switches). In this paper, we propose a data correction pipeline to generate ground-truth data more efficiently in this semi-automated scenario. Our method simplifies the trajectories from the tracking systems and let the annotator verify and correct the objects in the sampled keyframes. Once the objects in the keyframes are corrected, the bounding boxes in the other frames are obtained by interpolation.
Our method achieves substantial reduction in the number of frames requiring manual correction. In the MOT dataset, it reduces the number of frames by 30x while maintaining a HOTA score of 89.61\% . Moreover, it reduces the number of frames by a factor of 10x while achieving a HOTA score of 79.24\% in the SoccerNet dataset, and 85.79\% in the DanceTrack dataset. The project code and data are publicly released at \href{https://github.com/foreverYoungGitHub/trajectory-simplify-benchmark}{github/foreverYoungGitHub}.

\end{abstract}

\section{Introduction}

Object detection and tracking are core problems for video sport analytics \cite{sanford2020group,liu2021detecting,bhuiyan2020pose,cioppa2022soccernet, tora2017classification}, as well as other computer vision applications, such as surveillance systems \cite{Sultani_2018_CVPR,Revaud_2021_ICCV,Specker_2022_WACV} and autonomous vehicles \cite{luo2018fast,Kawasaki_2021_WACV,Greer_2022_WACV}. However, these tasks often require large datasets, and manual annotation in each frame is a time-consuming and expensive effort.


A practical semi-automated approach for data collection involves running off-the-shelf trackers, or an existing tracker for a specific application \cite{zhang2021fairmot,Yunhao2022StrongSORT,cao2022observation,zhang2022bytetrack}, and rely on a manual cleanup process. This process would correct for mistakes made by the tracking system, by considering errors in the detections (e.g. adding missing bounding boxes, adjusting them or deleting spurious detections), and errors in the association over time (e.g. track switches). 

Using tracking results as a prior for the annotation is often more efficient than starting from scratch, but verifying and cleaning the data for each frame still requires a significant effort. In this paper we explore ways to speed up the annotating process by subsampling the tracking data, selecting keyframes to be corrected, such that after correcting only the keyframes, the whole sequence can be obtained by interpolation.



A naive approach is to sample frames uniformly, but this approach is far from optimal, especially with complex trajectories, where some frames are more important than others (e.g. player changing direction). In this case, we can improve over uniform sampling by finding the most important keyframes to compress the trajectory, that is, treating it as a trajectory simplification problem. 
Several methods for trajectory simplification have been proposed to compress trajectory data in point-based applications \cite{douglas1973algorithms,chen2012fast,meratnia2004spatiotemporal,cao2017dots}, such as GPS data. These methods aim to reduce the number of points in a given trajectory, keeping a subset of the points that preserve important information about it. However, these methods were designed for point-tracking, and they are not optimized for bounding boxes (where the scale of an object over time is as important as its position). Another key difference for the problem at hand is that the input trajectories are noisy, and we are interested in preserving the quality of the trajectories \emph{after correction}. For this reason, the proposed method also takes the tracking confidence into consideration when selecting the keyframes.

In this paper, we introduce a scale-invariant trajectory simplification method for bounding box tracking, that shows potential to significantly reduce annotation times, while aiming to keep the tracking quality as close as possible to the quality obtained if all frames are corrected. Our method is shown in Fig \ref{fig:framework_eg}: given existing (noisy) trajectory data, it selects keyframes to be corrected. In order to find the optimal simplified trajectory, the keyframes are selected both from high-quality observation and outliers, such that the result minimizes the error metric for the recovered trajectories. We introduce a scale-invariant error metric to guide the trajectory simplification, that penalizes scale changes of the objects in the image. After the keyframes are manually corrected, the full trajectory is recovered by linear interpolation of the keyframes.
We perform a thorough evaluation on the MOT20\cite{dendorfer2020mot20}, SoccerNet\cite{cioppa2022soccernet} and DanceTrack\cite{sun2022dancetrack} datasets.
We consider two sets of experiments: (i) using tracking data from state-of-the-art detection (\cite{ge2021yolox}) and tracking (\cite{cao2022observation}, \cite{zhang2022bytetrack}) methods, and (ii) with synthetically corrupted ground truth tracks, that simulate common errors in trackers (e.g. bounding box jitter and track switches) and let us analyze the impact in performance as we vary the amount of noise in the tracking data. 
Our method is able to generate high-quality trajectory data even in scenarios where only $1/30$ of the frames are corrected in the MOT20 dataset, $1/5$ of the frames in the SoccerNet dataset and $1/10$ of the frames in the DanceTrack dataset,  outperforming existing trajectory simplification methods.


The key contributions of our work to the object tracking community are as follows:
\begin{itemize}
  \item We introduce a scale-invariant trajectory simplification method to speed up semi-automated data collection for object tracking in videos.
  \item We validate our proposed method on the MOT20, SoccerNet and DanceTrack datasets, that show improved performance compared to other trajectory simplification methods.
\end{itemize}

\section{Related Work}

\textbf{Video and Interactive Annotation}.
In recent years, the demand for video annotation tools has increased due to their vital role in training data for visual tasks. Interactive recurrent annotation framework introduced by Le \etal \cite{le2020toward}, and semi-automatic annotation method proposed by Ince \etal \cite{ince2021semi} have gained popularity. However, these methods still require checking all frames during annotation, leading to high annotation costs.

Existing video annotation tools, such as CVAT\cite{cvat} and VATIC\cite{vondrick2013efficiently}, offer a practical solution by using linear interpolation to generate bounding boxes and points for most frames, while partially annotating the key frames. However, these tools have limitations as they cannot integrate with existing semi-automatic annotation methods: the annotation process becomes time-consuming and costly by given automatically generated tracking trajectories, as the tools typically mark every frame as a keyframe, and annotators need to review and remove unnecessary annotations.



\textbf{Trajectory Simplification}.
Trajectory simplification presents one approach to automatically selecting useful keyframes in tracking trajectories. These methods have been utilized to compress trajectory data from a wide range of systems, including motion capture, touch screens, GPS, and IMUs. The goal of trajectory simplification is to reduce storage and computational resources, which is achieved by taking a sequence of size $N$ and obtaining a subsequence with $M$ points ($M \ll N$) that generates the minimum spatial distance error.

Dynamic programming (DP) \cite{bellman1961approximation} could be considered as the first algorithm for trajectory simplification, capable of guaranteeing to find the minimum error with O($N^3$) time complexity. DP was later improved in \cite{douglas1973algorithms}, with an approximate simplification method with error bound guarantee. This method minimizes the perpendicular Euclidean distance (PED), which is the shortest distance between points and their anchor segments. An extension of DP called TD-TR \cite{meratnia2004spatiotemporal} exploits the temporal dimension of trajectories and employs a new distance measure, called Synchronous Euclidean Distance (SED), that replaces the perpendicular distance used in DP when finding the split point with the maximum distance. SED considers the time information and uses it as the ratio to find the synchronized location of the points based on the linear interpolation. It calculates the distance between the actual point locations and their synchronized locations on the anchor segment.



A different approach involves constructing a Directed Acyclic Graph(DAG) and optimizing the trajectory by minimizing the integral of error metrics $\epsilon$. Optimization-based approaches \cite{cao2017dots, chen2012fast} compute the error metric for each corresponding timestamp and minimize the global integral error to improve performance. These methods use two integral errors, commonly known as integral square PED (ISPED) and integral square (ISSED).

It is worth noting that, unlike bounding box trajectories in vision problems, GPS or IMU trajectories are directly captured from the sensors and normally do not contain confidence scores. Therefore, current trajectory simplification methods do not rely on confidence information and cannot filter out low-quality observations. Additionally, these error functions present some issues when applied to bounding boxes, as they treat the two points that define a bounding box separately. This paper addresses these two issues.



\section{Problem Formulation}

We formulate the problem as follows: the inputs are noisy tracking observations $\textbf{B}_n$, consisting of bounding boxes $b_t$, scores $s_t$ for a set of frames $T_n$ for a given track id $n$.

\begin{equation}
\textbf{B}_n = \{b_t, s_t \}, t \in \textbf{T}_n
\end{equation}




The trajectory simplification task consists in finding the set $\textbf{B'}_n$ that subsamples the frames obtaining $\textbf{T'}_n$ ($|\textbf{T'}_n| \ll |\textbf{T}_n|$), aiming to preserve information on the original sequence. We call the ratio of $|\textbf{T}_n| / |\textbf{T'}_n|$  the \textit{compression ratio}. More precisely, the simplified trajectory $\textbf{B'}_n$ can be interpolated to recover the same number of frames of the original trajectory, obtaining $\textbf{B''}_n$. We can view the subsampled frames as keyframes for the trajectory. In this paper, we consider that $\textbf{B''}_n$ is recovered by linear interpolation of the simplified trajectory $\textbf{B'}_n$, and we compare $\textbf{B}_n$ and $\textbf{B''}$ with bounding box metrics (IoU) \cite{zheng2020distance} and tracking metrics (MOTA \cite{milan2016mot16}, HOTA\cite{luiten2020IJCV}).

\section{Proposed method}

\begin{figure}[t]
\vspace{-0.3cm}
\includegraphics[width=.45\textwidth, trim={0.23in 0 0.05in 0},clip]{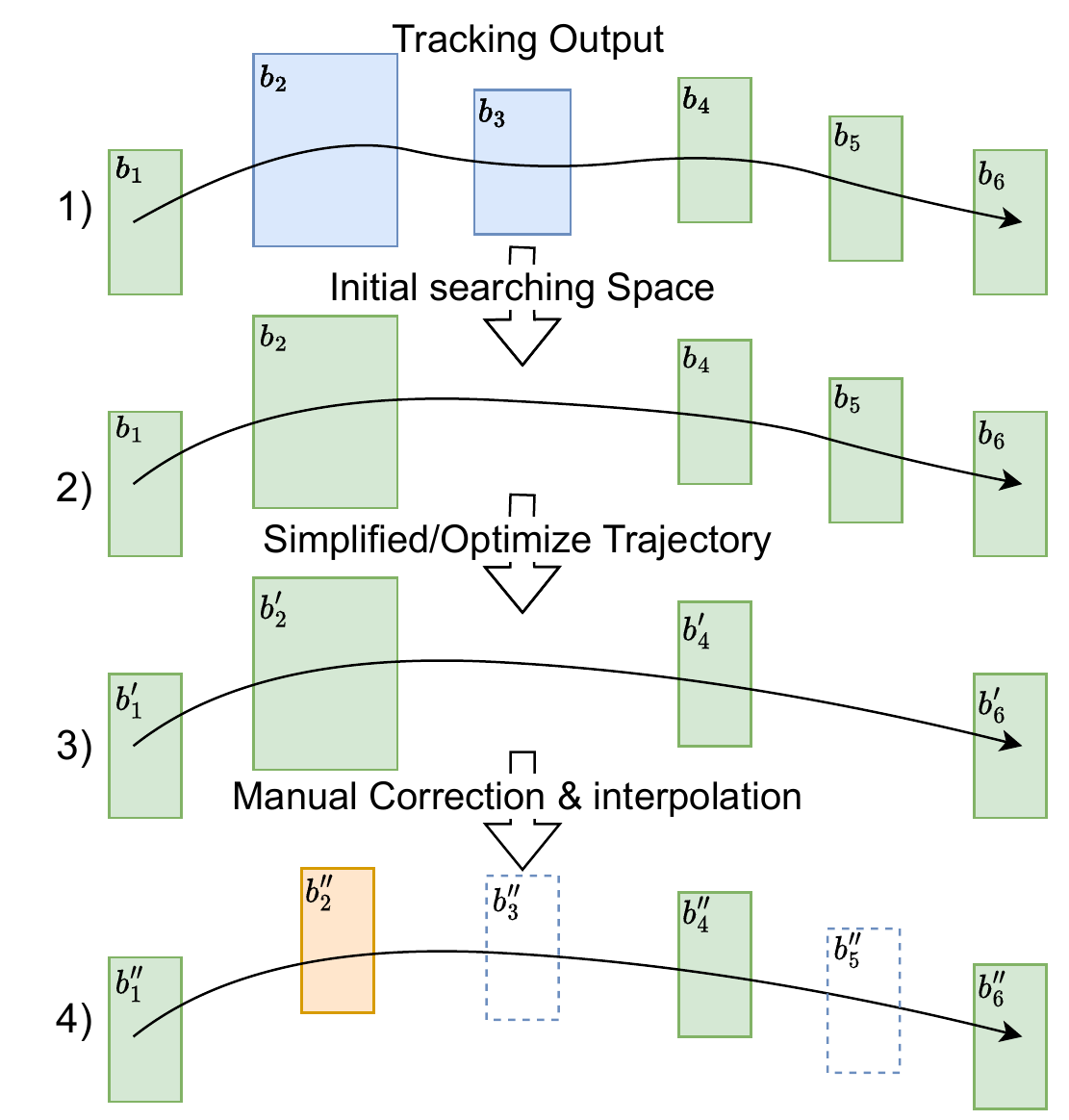}
   \caption{Illustration of the correction framework for one trajectory. Each line shows the trajectory in intermediate steps of the process: 1) the tracking trajectory, which is composed by high-quality observation ({\color{green}green}) and the low-quality observation ({\color{blue}blue}); 2) the initialized searching space (shown in \ref{sec:SearchingSpace}), which includes the high-quality observation and the outlier of low-quality observations ($b_2$); 3) the optimized trajectory in searching space by minimizing global error (shown in \ref{sec:simplification}); 4) verified ({\color{green}green}) and corrected ({\color{BurntOrange}orange}) the simplified trajectory, and recovered the whole trajectory by linear interpolation ({\color{blue}dashed blue})}
   \vspace{-0.5cm}
\label{fig:framework}
\end{figure}

We present a trajectory correction framework that utilizes tracking data as input and requires manual correction only for a small subset of this data to obtain the entire corrected sequence. Figure \ref{fig:framework} shows an overview of the framework: The proposed method categorizes observations into high-quality (\textcolor{green}{green}) and low-quality (\textcolor{blue}{blue}) subsets. Trajectories are then simplified by selecting a small set of keyframes from the full set using algorithms described in subsequent subsections. Only the selected keyframes require manual review or correction. Groundtruth trajectories are then interpolated from the cleaned simplified trajectories. 

\subsection{Initializing the Searching Space}
\label{sec:SearchingSpace}

To initialize the search space for tracking trajectories, a set of high-quality bounding boxes and outliers of low-quality bounding boxes are selected. The details of this step are presented in Algorithm \ref{alg:init}.

To filter out noisy observations and maintain necessary boxes to recover the trajectory, the high-quality bounding boxes are retained. Based on the assumption that a predicted bounding box $b_t$ with a higher score $s_t$ is closer to the ground truth box ${b_t''}$, the confidence scores $s_t$ of predicted bounding boxes are used to identify high-quality bounding boxes.

However, it is also important to include outliers of the low-quality bounding boxes to cover scenarios such as motion blur or irregular appearance which cannot be perfectly interpolated. For selection of the outliers, we draw inspiration from the Douglas-Peucker algorithm \cite{douglas1973algorithms}. For each anchor segment between high-confidence bounding boxes, we find the box $b_t$ causing the largest error $\epsilon$ with respect to a given tolerance threshold $\epsilon_{\text{th}}$. If the error $\epsilon$ is less than $\epsilon_{\text{th}}$, the approximation is accepted, and we only keep the two high-confidence bounding boxes while discarding the remaining boxes within the segment. If the error $\epsilon$ is greater than $\epsilon_{\text{th}}$, we split the segment into two sub-segments and add $b_t$ into the searching space $S$.

\begin{figure}[t]
\vspace{-0.3cm}
\includegraphics[width=.48\textwidth, trim={0in 0 0 0},clip]{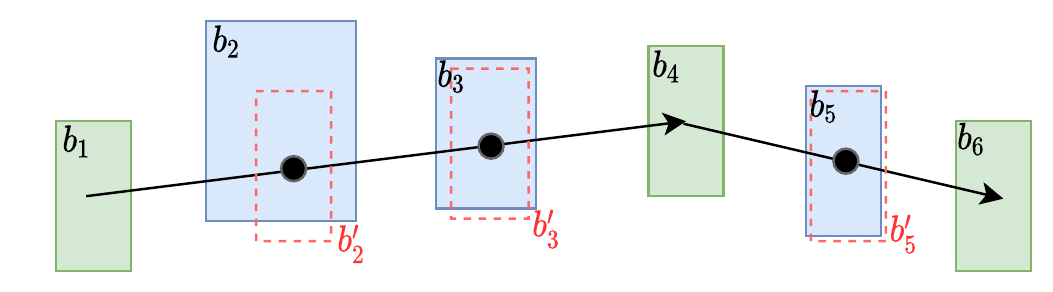}
   \caption{Illustration of $\text{Erf}_{IoU}$. The green boxes are the selected searching space, and it forms two tracking segments $b_1b_4$ and $b_4b_6$. The blue boxes are the predicted boxes not in the searching space, while the red boxes are the synchronized boxes for each tracking segment. The error metric for bounding boxes $\text{Erf}_{IoU}(b_t)$ calculates the IoU distance between the predicted bounding box $b_t$ and synchronized box $b_t'$.}
  \vspace{-0.40cm}
\label{fig:distance}
\end{figure}

\textbf{Scale-Invariant Error Metrics for Bounding Boxes}.
The central objective of our method is to generate highly compressed trajectories with low error. This is achieved by minimizing an error metric for the simplified trajectory. Error metrics proposed in the trajectory simplification literature are focused on point-based trajectories such as PED \cite{douglas1973algorithms,hershberger1992speeding} or SED \cite{meratnia2004spatiotemporal,chen2012fast}, which is not suitable for the bounding boxes trajectories. 

In the case of bounding boxes, the Intersection over Union (IoU) score is a commonly used metric for both training and evaluation. IoU is a scale-invariant metric that performs well in pinhole geometry. In this paper, we propose the synchronized IoU distance as the error metric to simplify the tracking trajectories. The synchronized IoU distance is calculated based on the IoU distance between the actual boxes ($b_t$) and their corresponding synchronized boxes ($b_t'$) on the anchor segment. An example of the synchronized IoU distance is shown in Figure \ref{fig:distance}. Here, synchronized IoU distance of box $b_5$ in between the anchor segment $b_4$ and $b_6$ is calculated as the IoU distance between $b_5$ and the corresponding synchronized box $b_5'$, which is linearly interpolated based on the time information.

To make the simplified trajectory more robust and reduce manual corrections, we incorporate the confidence score $s_t$ as weights into the synchronized IoU distance. The weighted IoU error metric is defined in Equation \ref{eq:error}.

\begin{align}
\text{Erf}_\text{IoU}(b_t, b_t', s_t) &= s_t \times (1 - IoU(b_t, b_t')) \label{eq:error}\\ 
&= s_t \times( 1 - \frac{I(b_t, b_t')}{U(b_t, b_t')} ) \nonumber 
\end{align}

where $I(b_t, b_t')$ means the intersection area of the actual predict bounding box $b_t$ and synchronized box $b_t'$, while the $U(b_t, b_t')$ means the union area of the $b_t$ and $b_t'$, $s_t$ means the confidence score of the predicted boxes $b_t$.


\begin{algorithm}[]
 \hspace*{\algorithmicindent} \textbf{Input} \\
 \hspace*{\algorithmicindent}    $B = \left \{ \left \{ b_1, s_1 \right \},...,\left \{ b_T, s_T \right \}\right \}$: boxes with score. \\
 \hspace*{\algorithmicindent}    $\Omega$: Confidence threshold. \\
 \hspace*{\algorithmicindent}    $\epsilon_{th}$: Error tolerance. \\
 \hspace*{\algorithmicindent}    Erf: Error metric. \\
 \hspace*{\algorithmicindent} \textbf{Output} \\
 \hspace*{\algorithmicindent}    $S$: Search spaces. \\
 \begin{algorithmic}[1]
 
 \State $S = \left [ \right ]$  
 
 \For{$i$ in range($T$)}
    \If {$s_{i} \ge \Omega$}
        Append index $i$ to $S$ 
    \EndIf
 \EndFor

  \For{$i = 0$ to $|S| - 1$}
    \State Get subset $\overline{B}$ of $B$ from index $S[i]$ to $S[i+1]$
    \State indices $L =$ Search($\overline{B}$)
    \State Insert indices $L$ to $S$ 
 \EndFor

%

 \State \Return sort($S$)
 
 \State
 
 \Procedure{Search}{$\overline{B},\epsilon,$Erf}
 \If {$len(\overline{B}) \le 2$}
     \Return $\left \{ \right \}$
 \EndIf
 \State $\epsilon, i = \max \text{Erf}(\overline{B})$
 \If {$\epsilon \le \epsilon_{th}$}
    \Return $\left \{ \right \}$
    
 \Else
    \State Split $\overline{B}$ to $\overline{B_1}$, $\overline{B_2}$ at index $i$
    \State \Return $\left \{  i \right \}$ + Search($\overline{B_1}$) + Search($\overline{B_2}$)
 \EndIf

 \EndProcedure

 \end{algorithmic}
 \caption{Initialize the Search Space}
 \label{alg:init}
\end{algorithm}

\subsection{Simplification by minimizing the integral error}
\label{sec:simplification}

\begin{figure}[t]
\vspace{-0.3cm}
\centering
\includegraphics[width=.18\textwidth, trim={0in 0 0 0},clip]{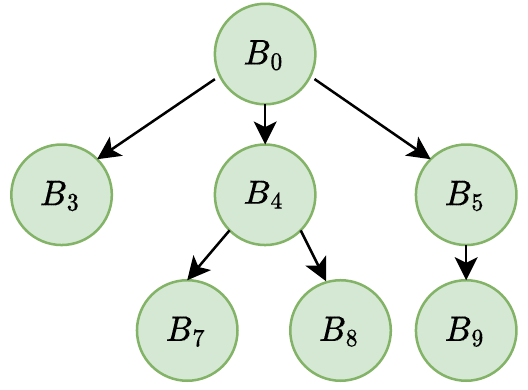}
  \vspace{-0.20cm}
   \caption{Illustration of constructed DAG, where the green nodes denote the boxes that are present in the search space $S$.}
  \vspace{-0.4cm}
\label{fig:dag}
\end{figure}

To reduce the number of nodes while minimizing the error metric, we utilize a Directed Acyclic Graph (DAG) that describes all potential simplified trajectories within a error tolerance threshold $\epsilon_{\text{th}}$. The DAG represents the observation nodes in the search space, with edges connecting any two vertices if the max error between them is less than the threshold $\epsilon_{\text{th}}$. In order to minimize the global integral error from the root vertex $V_0$ to $V_{T}$, each vertex $V_i$ stores the integral error from $V_0$ to $V_i$. The integral error of the node is obtained by integrating the previous integral error stored in parent vertices and the local errors between the current node and its connected parent node. In each step, the best parent node is selected to minimize the integral error. The DAG is constructed such that the integral error stored in $V_{T}$ is the minimum of the global integral error. An example of a constructed DAG is shown in Figure \ref{fig:dag}, where the initial searching space $S=\left \{ \textbf{B}_0, \textbf{B}_3, \textbf{B}_4, \textbf{B}_5, \textbf{B}_7, \textbf{B}_8, \textbf{B}_9 \right \}$. We begin by checking if the root box $\textbf{B}_0$ can connect to other boxes. Here, since the max $\text{Erf}_\text{IoU}$ between $\textbf{B}_3, \textbf{B}_4, \textbf{B}_5$ and $\textbf{B}_0$ is less than $\epsilon_{\text{th}}$, we connect them to the root vertex. We do not build the connection between $\textbf{B}_0$ and $\textbf{B}_7$ since the max $\text{Erf}_\text{IoU}(\textbf{B}_0 \to \textbf{B}_7) > \epsilon_{\text{th}}$. We stop checking the remaining boxes once max $\text{Erf}_\text{IoU}(\textbf{B}_0 \to \textbf{B}_8) > \epsilon_{\text{th}}$. We then proceed to check the connections between $\textbf{B}_7, \textbf{B}_8, \textbf{B}_9$ and the current parent vertices $\textbf{B}_3, \textbf{B}_4, \textbf{B}_5$, until we construct the complete DAG.


To obtain the simplified trajectory, we follow the unique path from the last vertex $B_T$ to the first vertex $B_0$ in reverse order. For instance, the final simplified trajectories in Fig \ref{fig:dag} is $\left \{ \textbf{B}_0, \textbf{B}_5, \textbf{B}_9 \right \}$. The algorithm for minimizing the error metric is provided in Algorithm \ref{alg:minimize-epsilon}. Notably, this approach is similar to previous work such as \cite{cao2017dots,chen2012fast}. By utilizing the DAG and minimizing the integral error, our approach can effectively reduce the number of nodes in the search space while maintaining a high level of accuracy.

\textbf{General Integration (GI) Function.}
The integral error in each layer integrates local errors between the current node and parents and the integral error in the previous path stored at parents node. To form a general integral function, we extend previous integral functions, such as ISPED\cite{chen2012fast} and ISSED\cite{cao2017dots}, by using the $n$-norm to combine errors, which is defined in the Equation \ref{eq:gi}.




\begin{align}
GI_{0 \to 1} &= \left | \left [ \epsilon_0, \epsilon_1 \right ] \right |_n \nonumber \\
GI_{0 \to c} &= \left | \left [ GI_{0 \to p}, \epsilon_{p+1}, ..., \epsilon_c \right ] \right |_n \label{eq:gi} \\
&= \left | \left [ \epsilon_0, ..., \epsilon_c \right ] \right |_n \nonumber
\end{align}

where $GI_{0 \to 1}$ is the integral error from the root vertex (index 0) to one of its child vertices (index 1), while $\epsilon_{0}$ is the error at index 0. $GI_{0 \to c}$ denotes the integral error from the root vertex (index 0) to the child vertex (index c), and $GI_{0 \to p}$ represents the previous integral error from the root to parent vertex (index p) stored in the parent node.


The equation states that the $n$-norm of the error from the root vertex $V_0$ to $V_c$ is equal to the $n$-norm of the previous integral error and the set of errors between the parent and child nodes. This means that the GI between the parent and leaf nodes is equal to the global integral function. By minimizing the GI in each layer, we can minimize the global integral error.

The value of $n$ in the $n$-norm balances the mean and max of the error samples. When $n=1$, it represents the sum of all the error metrics in each timestamp. When $n=2$, it represents the integral square of the error metrics in each subseries. When $n$ is infinite, the integral error represents the maximum error in each subseries.

\begin{algorithm}[]
 \hspace*{\algorithmicindent} \textbf{Input} \\
 \hspace*{\algorithmicindent}    $B = \left \{ \left \{ b_1, s_1 \right \},...,\left \{ b_T, s_T \right \}\right \}$: boxes with score. \\
 \hspace*{\algorithmicindent}    $S = \left \{i_1,...,i_M\right \}$: Search space. \\
 \hspace*{\algorithmicindent}    $\epsilon_{th}$: Error tolerance. \\
 \hspace*{\algorithmicindent}    Erf: Error metric. \\
 \hspace*{\algorithmicindent}    GI: Integral function. \\
 \hspace*{\algorithmicindent} \textbf{Output} \\
 \hspace*{\algorithmicindent}    $R$: Result for the simplified indices. \\

 \begin{algorithmic}[1]
 \State //visit status $V$: unvisited, current leaf, current parent, visited
 
 \State set $V$ to visited for all the nodes in search space $S$
 \State set $V_0$ as the root node with status current parent
 \State set $E=[0, \text{inf}, ..., \text{inf}]$ to store the integral errors $GI_{0 \to t}$ for all $B$
 \State set $P$ for parent node index of all $B$
 
 \While {$V_{end}$ not visit}
    \For{$i_c$ in unvisited nodes}
        \For{$i_p$ in current parent nodes}
            \State Get subset $\overline{B}$ of $B$ from index $i_p$ to $i_c$
            \State $\epsilon_{\overline{B}} = \text{Erf}(\overline{B})$ \Comment{$\epsilon_{\overline{B}}$ is a list of errors}
            \State $\epsilon = \max(\epsilon_{\overline{B}_S})$
            \State $\text{GI}_{0 \to i_c} = \text{GI}(E_{i_p}, \epsilon_{\overline{B}})$
            \If {$\epsilon < \epsilon_{th}$ and $\text{GI}_{0 \to i_c}< E_{i_c}$}
                \State set $V_{i_c}$ is current leaf
                \State set $E_{i_c} = \text{GI}_{0 \to i_c}$ and $P_{i_c} = i_p$
            \ElsIf{$\epsilon > 2 \epsilon_{th}$}
                break to while
            \EndIf
        \EndFor
    \EndFor
    \State set $V$ of node in current parents to visited
    \State set $V$ of node in current child to current parents
 \EndWhile
 
 \State \Return ravel through the unique path from $P_{end} \to P_{0}$ for parent node index of all $B_S$
 \end{algorithmic}
 \caption{Minimizing the integral error}
 \label{alg:minimize-epsilon}
\end{algorithm}

\section{Experimental Protocol}



We conducted experiments to evaluate the proposed trajectory correction algorithm on both real tracking data and synthetic data with various levels of noise. 
In both cases, we utilize a simulated correction pipeline, by matching the predicted bounding boxes on the selected keyframes with the ground truth data. Specifically, this experimental protocol assumes that any frame selected for manual review is perfectly corrected. After correcting the keyframes, we interpolate the trajectory and report metrics that compare the interpolated corrected trajectory with the ground truth. It should be noted that we only used linear interpolation to ensure compatibility with existing annotation toolboxes.

For these experiments, unless explicitly stated, the confidence threshold $\Omega$ was selected based on the top 10 percentile of each trajectory. The error metric used was the IoU distance, while the $n$ value was set to 1 in the general integral function. As there are no similar approaches for this task, we compared our method with uniform sampling and the SOTA DP-based\cite{meratnia2004spatiotemporal} and DAG-based\cite{chen2012fast} point trajectory simplification methods. For the latter, we simplified the bounding box trajectories using the top-left and bottom-right points of the box. 

\textbf{Datasets and Evaluation Metric.}
We evaluate our proposed method on MOT20 \cite{dendorfer2020mot20}, SoccerNet \cite{cioppa2022soccernet} and DanceTrack \cite{sun2022dancetrack} datasets. The MOT20 dataset \cite{dendorfer2020mot20} consists of multi-person tracking data for pedestrians, with the scale of the bounding box being relatively consistent in the full track. SoccerNet \cite{cioppa2022soccernet} is a tracking dataset consisting of multiple objects, including players, the ball, and referees, in soccer broadcast videos, that may involve camera movement. In this paper, we consider only tracking of the \emph{person} class in the evaluation. The DanceTrack \cite{sun2022dancetrack} dataset consists of multi-person tracking in dance videos. In DanceTrack, the bounding boxes vary significantly in the nearby frames and the entire sequence. To evaluate the accuracy of the interpolated corrected trajectory, we adopt IoU scores in detection metrics and MOTA \cite{milan2016mot16} and HOTA \cite{luiten2020IJCV} in tracking metrics.


\textbf{Input tracking data.}
For the real tracking data, we apply YoloX \cite{ge2021yolox} pretrained on CrowdHuman \cite{shao2018crowdhuman} and MOT20
\cite{dendorfer2020mot20} datasets to extract bounding boxes in each frame, and we use the combination of the ocsort \cite{cao2022observation} and byte \cite{zhang2022bytetrack} algorithms to track the bounding boxes without any re-identification features. We match the actual trajectory with the ground truth trajectory by Hungarian assignment in each frame to associate the bounding boxes with the corresponding ground truth data. Each actual trajectory is then simplified based on our proposed method, generating a simplified trajectory that is ready for data correction. To mimic the data correction process, we assign the ground truth bounding box and track IDs back to the keyframes in the simplified trajectory to fix the bounding box jitters and ID switches. In cases where the first or last objects in the ground truth trajectory do not have a corresponding match in the tracking data, we add these missing objects to the corrected trajectory in order to align it with the ground truth trajectory and facilitate meaningful comparison.


\begin{figure}[]
\vspace{-0.3cm}
 \begin{subfigure}[b]{0.161\textwidth}
     \centering
     \includegraphics[width=\textwidth, trim={0.1in 0.17in 0 0},clip]{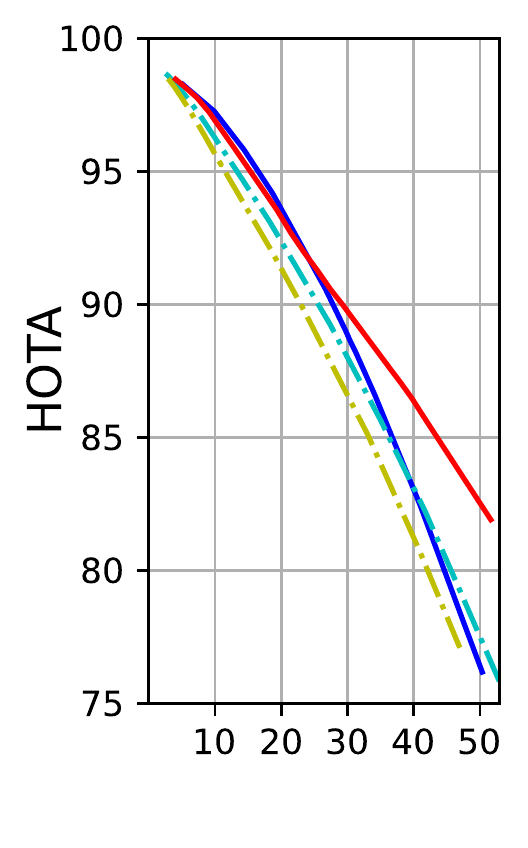}
     \caption{MOT20}
     \label{fig:mot_actual_result}
 \end{subfigure}
 \begin{subfigure}[b]{0.146\textwidth}
     \centering
     \includegraphics[width=\textwidth, trim={0.1in 0.1in 0 0},clip]{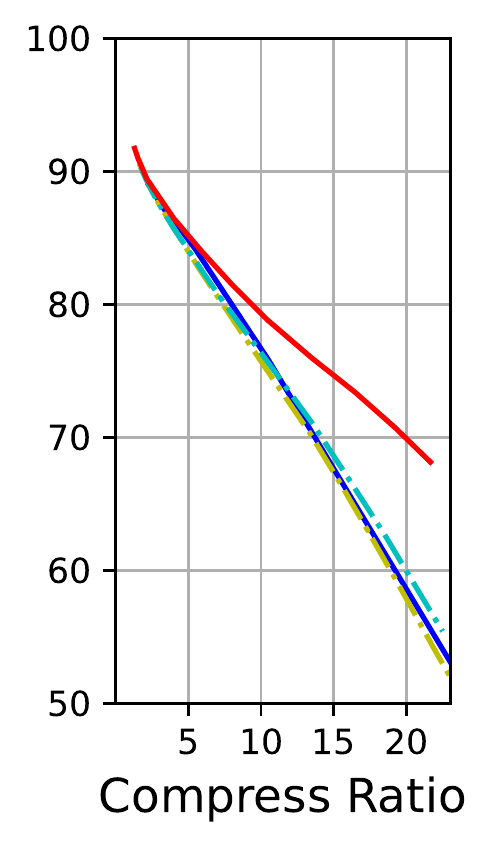}
     \caption{SoccerNet}
     \label{fig:soccer_actual_result}
 \end{subfigure}
 \begin{subfigure}[b]{0.146\textwidth}
     \centering
     \includegraphics[width=\textwidth, trim={0.1in 0.2in 0 0},clip]{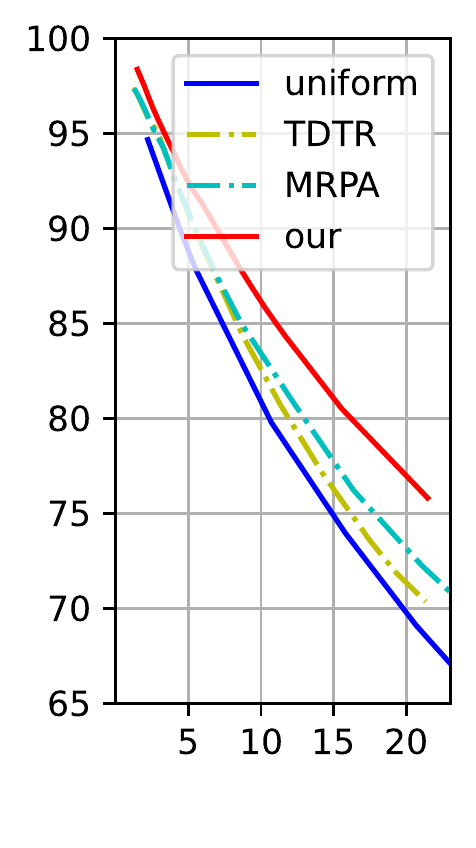}
     \caption{DanceTrack}
     \label{fig:dance_actual_result}
 \end{subfigure}

  \vspace{-0.2cm}
  \caption{The HOTA score for trajectory correction as we vary the compression rate. The uniform sampling, TDTR \cite{meratnia2004spatiotemporal}, MRPA \cite{chen2012fast}, and our proposed method are represented by the blue, yellow, cyan, and red curves, respectively. A higher accuracy score indicates better correction results.}
  \vspace{-0.5cm}
\label{fig:actual_result}
\end{figure}


\textbf{Synthetic data.}
To investigate the algorithm's sensitivity to different levels of noise in the input tracking data, we perform additional experiments using synthetic data. We corrupt the ground truth trajectories with two common tracking mistakes: bounding box jitter and track switches. We simulate bounding box jitter by perturbing the bounding box positions (center and scale) with Gaussian noise. For track switches, we switch the tracking IDs of bounding boxes with high overlap (IoU $> 0.5$) with a probability $p$. A probability of 0 indicates that the input trajectory is noise-free and therefore the process is equivalent to simplifying ground truth trajectory. Following the same procedure used for the actual tracking data, we apply the correction methods to select the keyframes and correct them based on the matched ground truth data.

\section{Result and Discussion}



\subsection{Tracking trajectory correction}
\label{sec:real_correction}


To validate the performance of our algorithm on real tracking data, we experiment with different compression ratios on the three selected datasets. The compression ratio indicates how much we compressed the trajectory (e.g. a ratio of 10 indicate that 1/10th of the frames were selected for correction). We report the results in terms of HOTA scores, which are shown in Figure \ref{fig:actual_result}. Our algorithm outperforms existing methods on both datasets. In general, we observe that the higher the compression ratio, the greater the improvement gained from our algorithm.


\begin{table}[]
\resizebox{\columnwidth}{!}{
\centering
\begin{tabular}{ccccc}
\hline \hline
MOT20 \cite{dendorfer2020mot20}      & IoU(mean)            & IoU(min)             & MOTA                 & HOTA                 \\ \hline
Raw Tracking         & -               & -              &  93.28\%               &  75.14\%             \\
Uniform(x30)         & 88.58\%               & 60.52\%              &     98.14\%         &   88.59\%     \\ 
TD-TR\cite{meratnia2004spatiotemporal}(x30)         & 86.01\%               & 59.47\%              &     96.89\%         &   82.23\%     \\ 
MRPA\cite{chen2012fast}(x30)         & 87.44\%	           & 62.57\%              &     98.58\%         &   85.12\%     \\ 
ours(x30)            & \textbf{91.37\%}     & \textbf{70.28\%}     &    \textbf{99.36\%}        & \textbf{89.61\%}        \\ \hline \hline
SoccerNet \cite{cioppa2022soccernet} & IoU(mean)            & IoU(min)             & MOTA                 & HOTA                 \\ \hline
Raw Tracking         & -               & -              &  49.22\%               &  49.55\%             \\
Uniform(x10)         & 79.81\%              & 12.55\%              &       84.07\%        &   75.77\%                   \\ 
TD-TR\cite{meratnia2004spatiotemporal}(x10)         & 79.04\%               & 14.34\%              &     85.97\%        &     74.18\%    \\  
MRPA\cite{chen2012fast}(x10)         & 80.08\%	           & 15.09\%              &      87.66\%        &   75.5\%     \\
ours(x10)           & \textbf{82.96\%}     & \textbf{16.2\%}     &       \textbf{89.7\%}      &  \textbf{79.24\%}  \\ \hline \hline
DanceTrack \cite{sun2022dancetrack} & IoU(mean)            & IoU(min)             & MOTA                 & HOTA                 \\ \hline
Raw Tracking         & -               & -              &  91.33\%               &  58.7\%             \\
Uniform(x10)         & 83.54\%              & 31.75\%              &      93.78\%         &   79.78\%                    \\
TD-TR\cite{meratnia2004spatiotemporal}(x10)         & 83.4\%               & 37.58\%              &     94.95\%         &   80.72\%     \\ 
MRPA\cite{chen2012fast}(x10)         & 84.31\%	           & 40.86\%              &     95.09\%         &   80.9\%     \\ 
ours(x10)           & \textbf{87.54\%}     & \textbf{46.58\%}     &       \textbf{98.46\%}      &  \textbf{85.79}\%           
\end{tabular}
}
\vspace{-0.3cm}
\caption{The data correction accuracy for real trajectory in MOT20(upper) and DanceTrack(lower) datasets. The x30 and x10 in the scope means the compression rate is 30 times and 10 times respectively. The higher accuracy score means the better result we get.}
\vspace{-0.5cm}
\label{tab:actual_result}
\end{table}

Figure \ref{fig:mot_actual_result} revealed an intriguing pattern on the MOT20 dataset: at low compression rates, trajectory simplification methods may underperform compared to the straightforward uniform sampling. For instance, when the compression rate is less than 20x, our algorithm performs similarly or slightly worse than uniform sampling. Uniform sampling works well in this scenario due to the MOT20 trajectories having linear motion within the short time window \cite{sun2022dancetrack}, while trajectory simplification methods tend to prioritize covering outliers caused by ID switches instead of reducing the frame gap, which could decrease detection accuracy (DetA). However, when the time window increases, non-linear motions become more prevalent within the trajectories, which is where our method excels by increasing the compression rate. Conversely, in the DanceTrack dataset (Figure \ref{fig:dance_actual_result}), all trajectory simplification methods outperform uniform sampling with all compression rates by selecting keyframes that capture the trajectories' significant variations resulting from large and non-linear motions.


Another important observation from Figure \ref{fig:actual_result} is that reducing the compression rate is critical to obtaining high-quality ground truth trajectories for complex motions and noisy tracking trajectories. This is due to the fact that, for the same compression rate, the accuracy of the corrected trajectory in SoccerNet is lower than that in the less complex much cleaner MOT20 dataset. For instance, when our method is used to sample trajectories 20 times and correct them, the mean IoU and HOTA scores for the fully recovered trajectories in MOT20 are 93.93\% and 93.16\%, respectively, while the mean IoU and HOTA scores decrease to 75.48\% and 70.76\% in SoccerNet. It should be noted that a visual shift in bounding boxes may be observed when the IoU scores are lower than 90\%.


Table \ref{tab:actual_result} presents the accuracy scores of the raw tracking trajectories and the corrected trajectories. The simulated correction pipeline is effective in significantly boosting the HOTA scores since all id switches are corrected in the recovered trajectory. For instance, in the DanceTrack dataset, the HOTA score only achieves 58.7\% in the raw tracking trajectories due to the low association accuracy (41.91\%). However, our proposed correction method is capable of increasing the HOTA score to 85.79\%. 

The results in Table \ref{tab:actual_result} demonstrate that the classic point-based trajectory simplification methods, TDTR \cite{meratnia2004spatiotemporal} and MRPA \cite{chen2012fast}, perform similarly or slightly worse than uniform sampling in both datasets when the mean IoU metric is considered. This suggests that these methods struggle to identify keyframes for the visual bounding box trajectories. In contrast, our proposed model achieves substantially higher mean IoU scores in all datasets.
Moreover, although trajectory simplification methods may slightly underperform compared to uniform sampling for the entire trajectories, they are more effective in covering the outlier cases based on the min IoU metric. It is important to note that our proposed method has a more significant impact on the outlier cases, as compared to uniform sampling. In particular, using our method, the min IoU scores for the corrected trajectories increase by 9.76\% and 14.83\% for the MOT20 and DanceTrack datasets, respectively, demonstrating its effectiveness in handling challenging tracking scenarios.

\subsection{Algorithm Modules Analysis}
\label{sec:real_ablation}

To understand our algorithm better, we carried out controlled experiments to examine how each component affects performance. For all experiments, we use the same settings and real tracking trajectory data, except for specified changes to the settings or component(s). The relative results are shown in Table \ref{tab:analysis}.

\begin{table}[]
\resizebox{\columnwidth}{!}{
\begin{tabular}{cccc|cc}
HighConf   & Outlier    & min-$\epsilon$ & Dist & $IoU_{mean}$ & $IoU_{min}$ \\ \hline
           & \checkmark &                & SED  & 83.4\%       & 37.58\%      \\
           &            & \checkmark     & SED  & 84.31\%      & 40.86\%     \\
           & \checkmark &                & IoU  & 86.03\%      & 47.18\%      \\
           &            & \checkmark     & IoU  & 87.33\%      & 45.91\%     \\
\checkmark &            & \checkmark     & IoU  & 86.81\%      & 39.83\%      \\
\checkmark & \checkmark & \checkmark     & IoU  & 87.54\%      & 46.58\%     \\
\checkmark & \checkmark & \checkmark     & DIoU & 87.74\%      & 46.87\%    
\end{tabular}
}
\vspace{-0.3cm}
\caption{Ablation studies on different modules in our proposed method, which were evaluated on the actual track generated from the DanceTrack dataset at an approximate compression rate of 10x.} 
\vspace{-0.5cm}
\label{tab:analysis}
\end{table}




\textbf{Using scale-invariant error metrics, such as IoU, is crucial for accurate visual tracking annotation.} Table \ref{tab:analysis} shows that adopting the IoU as the error metric leads to the most significant improvement, boosting mean IoU scores by 3\%. Further improvements in accuracy can be achieved by using the DIoU \cite{zheng2020distance} and CIoU \cite{zheng2021ciou} distance metrics, which increase mean IoU scores by 0.2\%.

\textbf{Combining high-quality and outlier boxes for initializing the searching space produces the best results.} When optimizing the integral error for only high-quality bounding boxes, useful information is removed, resulting in a min IoU score of 39.83\%. However, incorporating all bounding box trajectories can bias the simplified trajectories with noisy low-quality bounding boxes. By combining high-quality and outlier bounding boxes, our proposed method is able to filter noise while retaining necessary trajectory information, achieving a mean IoU score of 87.74\% and a min IoU score of 46\%.

\subsection{Trajectory Noise Impact Analysis}
\label{sec:synthetic_ablation}

We now consider the results on the synthetic dataset, in order to quantify the impact on performance of bounding box jitter and track id switches. 

\begin{figure}[]
\vspace{-0.3cm}
    \begin{subfigure}[b]{.48\textwidth}
     \centering
     \addtolength{\tabcolsep}{-5pt}    
     \begin{tabular}{c c}
      \rotatebox[origin=c]{90}{\small{(a) Noisy Detection}} & \includegraphics[valign=m,width=0.9\textwidth, trim={0.1in 0.1in 0.1in 0.1in},clip]{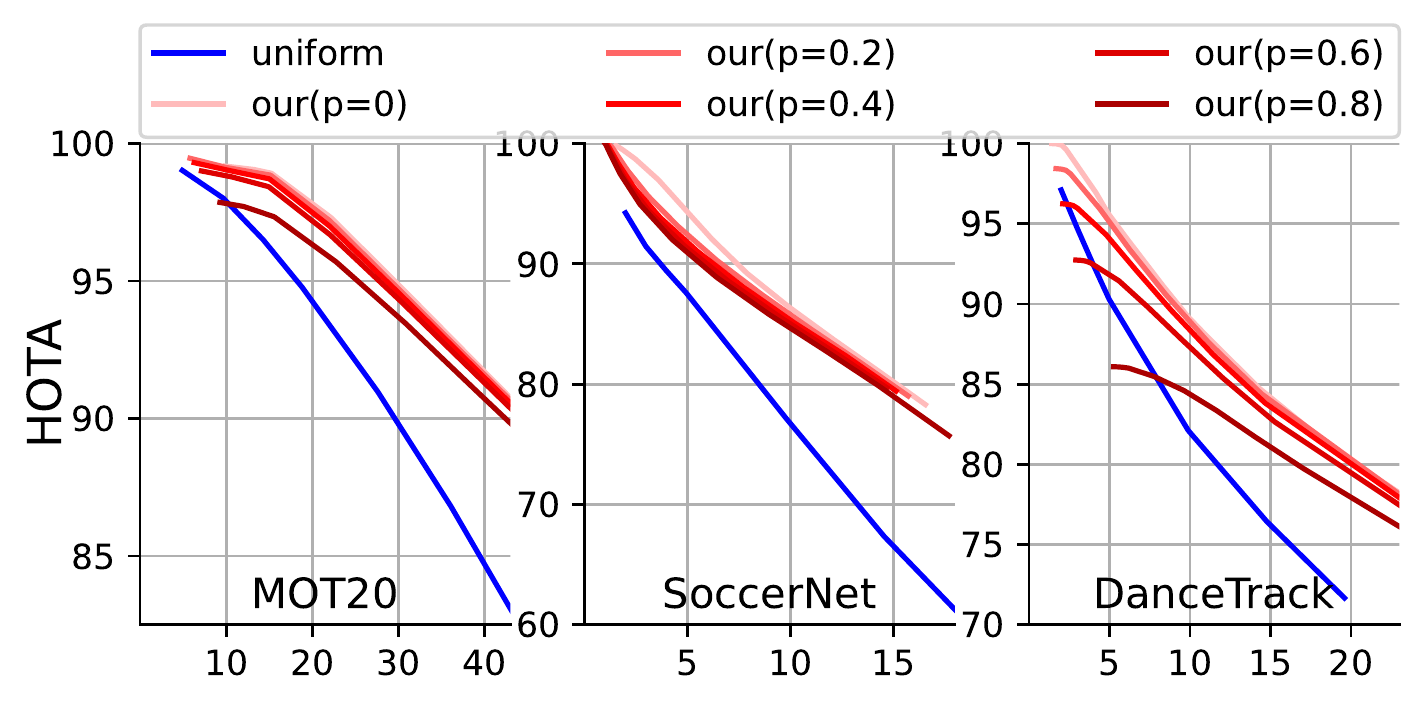}
    \end{tabular}
    \addtolength{\tabcolsep}{1pt}
   \end{subfigure}
   
   \vspace{0.2cm}
   
   \begin{subfigure}[b]{.48\textwidth}
     \centering
     \addtolength{\tabcolsep}{-5pt}    
     \begin{tabular}{c c}
      \rotatebox[origin=c]{90}{\small{(b) Noisy ID Switch}} & \includegraphics[valign=m,width=0.9\textwidth, trim={0.1in 0.1in 0.1in 0.1in},clip]{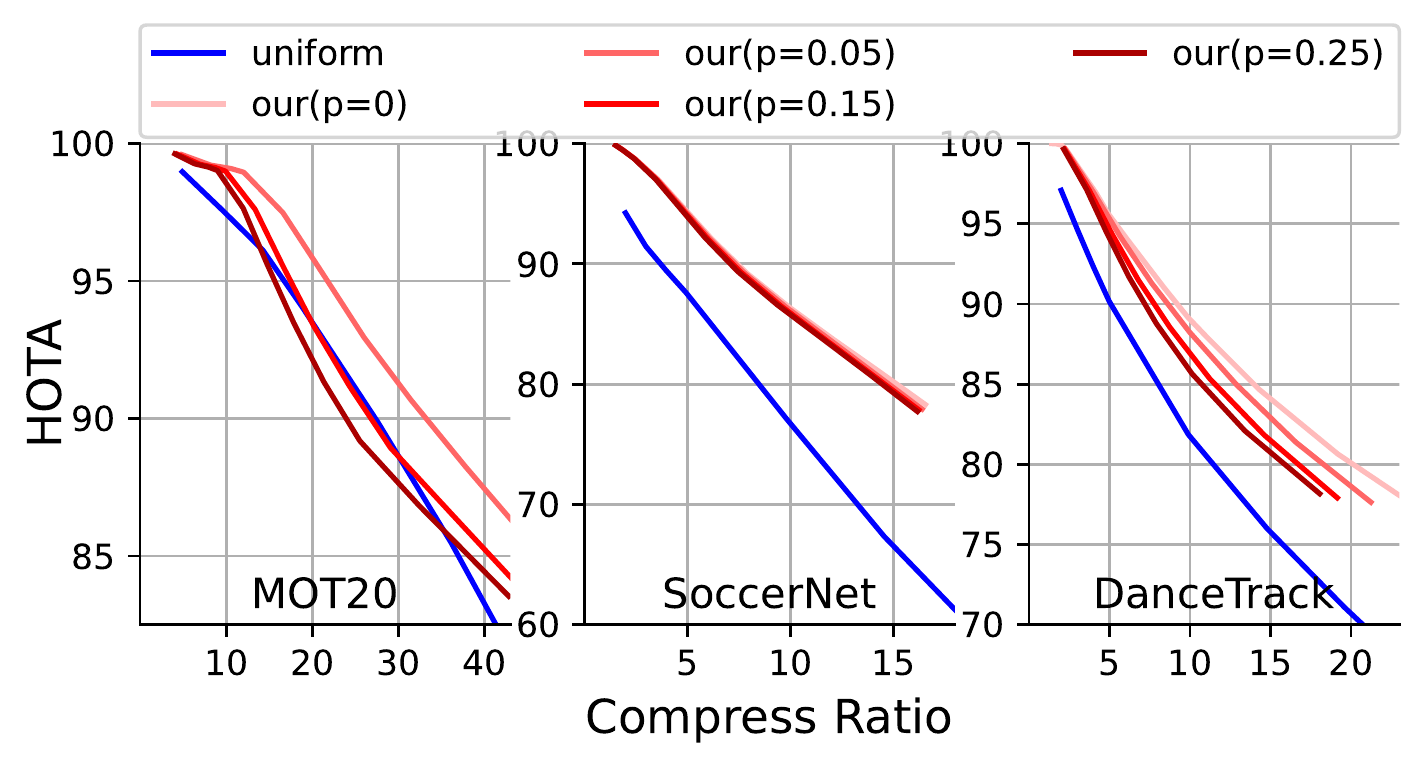}
    \end{tabular}
    \addtolength{\tabcolsep}{1pt}
   \end{subfigure}

  \vspace{-0.3cm}
  \caption{The HOTA score for trajectory correction with noisy bounding box (a) and noisy id switches (b). The blue curves represent the scores for uniform sampling, whereas the red curves correspond to the results obtained from our proposed methods. The label on the curves includes the noisy probability $p$, with darker lines indicating higher levels of detection noise / id switches in the trajectories.}
  \vspace{-0.5cm}
\label{fig:noise_result}
\end{figure}

\textbf{Correction of trajectories with noisy detection}. The HOTA scores of the corrected trajectories are shown in Fig \ref{fig:noise_result}a. Here, we only use uniform sampling as a baseline reference since it is not impacted by the bounding box noise, and therefore only one curve is plotted for uniform sampling.

In Figure \ref{fig:noise_result}a, we observed that our proposed algorithm outperforms uniform sampling in general. The HOTA scores generated by our algorithm closely match the simplified ground truth trajectory ($p=0$) when the noise probability is less than 0.4. This demonstrates that our algorithm effectively filters out the noisy bounding boxes and retains the most informative keyframes. However, the accuracy score notably decreases when the noisy jitter rate surpasses 0.4, especially at low compression rates. This is because nearly every bounding box experiences shifts when the noise probability is extremely high. In such cases, as we minimize the integral error by sampling more data, we inadvertently incorporate more noisy bounding box jitters. Thus, the proposed algorithm is highly sensitive to high levels of bounding box jitter.

\textbf{Correction of trajectories with track id switches}. The HOTA scores of the corrected trajectories are shown in Fig \ref{fig:noise_result}b. Our proposed algorithm consistently outperforms uniform sampling in all sampling rates and id switch noise rates. Moreover, the algorithm is able to capture errors that occur when id switches happen, resulting in HOTA scores that are close to the simplified ground truth trajectory when the noise rate is low. However, when the noise probability is high, such as in the case of the MOT20 dataset with a sampling rate of 20x, there is a concave curve in the HOTA scores of our algorithm. That is because there are many id switches that occur frequently back and forth in the adjacent frames with high noise probability. In this case, the uniform sampling directly skips some of the id switches segments, while our algorithm captures these id switches densely as they happen, sacrificing the accuracy of the rest of the trajectory, which supports our assumption in Section \ref{sec:real_correction}. It should be noted that the id switches were simulated at higher frequencies than what is typically observed in real tracking data, as the objective of this section was to evaluate the algorithm's performance under varied levels of noise.


It is worth noting that although we have added significant amounts of detection and tracking noise to generate the synthetic dataset, the trajectories remain mostly complete without any missing frames. This is in contrast to real data, as illustrated by the example of SoccerNet in Fig \ref{fig:soccer_actual_result}), the HOTA score after correcting all frames in the tracking trajectories is only 92.22\% due to the incompleteness of the generated trajectories by the pre-trained model. On the other hand, based on the result shown in Fig \ref{fig:noise_result}, if we can provide acceptable trajectories for SoccerNet generated by the fine-tuned model, our method can achieve 90.21\% HOTA by correcting only 3 frames per second, and 94.75\% HOTA by correcting only 5 frames per second. In comparison, the uniform sample method requires correcting half of the frames to achieve a HOTA score of 94.23\%.

\section{Limitation}

Our experimental results have identified certain limitations of the proposed method. It has been observed that if the tracked motion predominantly follows a linear pattern, the proposed method may not demonstrate an improvement over uniform sampling. As observed in MOT20 dataset, where the proposed method only outperforms uniform sampling at compression rates exceeding 20x. Additionally, limited to the existing annotation tools, the use of linear interpolation between keyframes imposes a constraint on the extent to which complex trajectories can be compressed. As observed in the DanceTrack dataset, although the proposed method exhibits significant improvements over other methods, when the compression rate is over 10x, the IoU drops below 90\% which leading to visible errors.




\section{Conclusion}

In this paper, we introduced a scale-invariant trajectory simplification method to minimize the annotation cost for semi-automated bounding box trajectory collection. The proposed method selects keyframes for each object in the video, such that only the keyframes needs manual review and correction. The experiments conducted on three popular tracking datasets demonstrate that our method can generate high-quality annotation data while requiring correction of significantly fewer frames. We also conducted ablation studies that showed that using a scale-invariant error metric is crucial for the task of simplifying bounding-box tracking data. Future work can consider extending these formulations to other vision trajectory tasks, such as pose tracking. 

{\small
\bibliographystyle{ieee_fullname}
\bibliography{egpaper}
}

\end{document}


\title{Supplementary Material for the paper #1408 - A Scale-Invariant Trajectory Simplification Method for Efficient Data Collection in Videos}

\author{First Author\\
Institution1\\
Institution1 address\\
{\tt\small firstauthor@i1.org}
\and
Second Author\\
Institution2\\
First line of institution2 address\\
{\tt\small secondauthor@i2.org}
}

\maketitle


\section{Visualizing tracks with different IoUs}

\begin{figure*}[]
\centering
\begin{subfigure}[b]{0.86\textwidth}
     \centering
     \includegraphics[width=\textwidth,trim={0 0 0 0},clip]{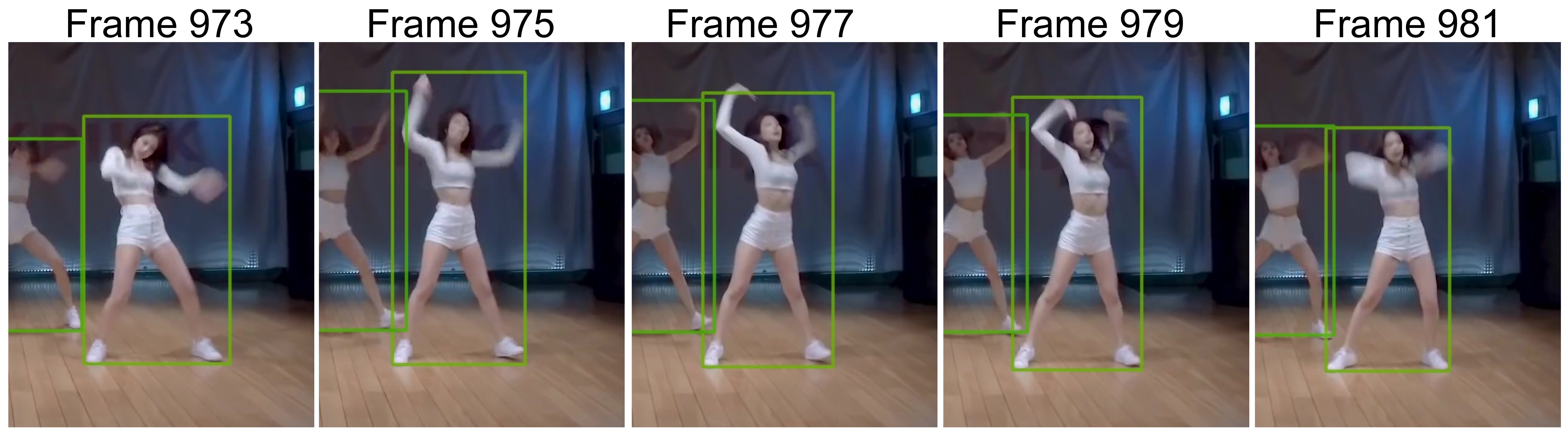}
     \caption{$IoU_{mean}=0.95$, obtained with compression rate 3.7x}
     \label{fig:demo_iou_1}
 \end{subfigure}
 \begin{subfigure}[b]{0.85\textwidth}
     \centering
     \includegraphics[width=\textwidth,trim={0 0 0 0},clip]{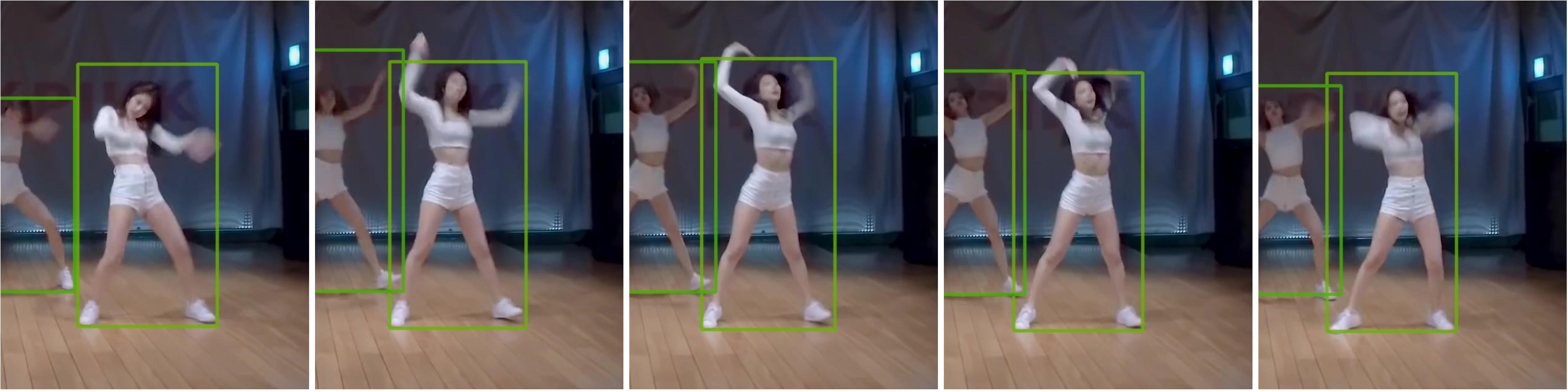}
     \caption{$IoU_{mean}=0.90$, obtained with compression rate 8.5x}
     \label{fig:demo_iou_2}
 \end{subfigure}
 \begin{subfigure}[b]{0.85\textwidth}
     \centering
     \includegraphics[width=\textwidth,trim={0 0 0 0},clip]{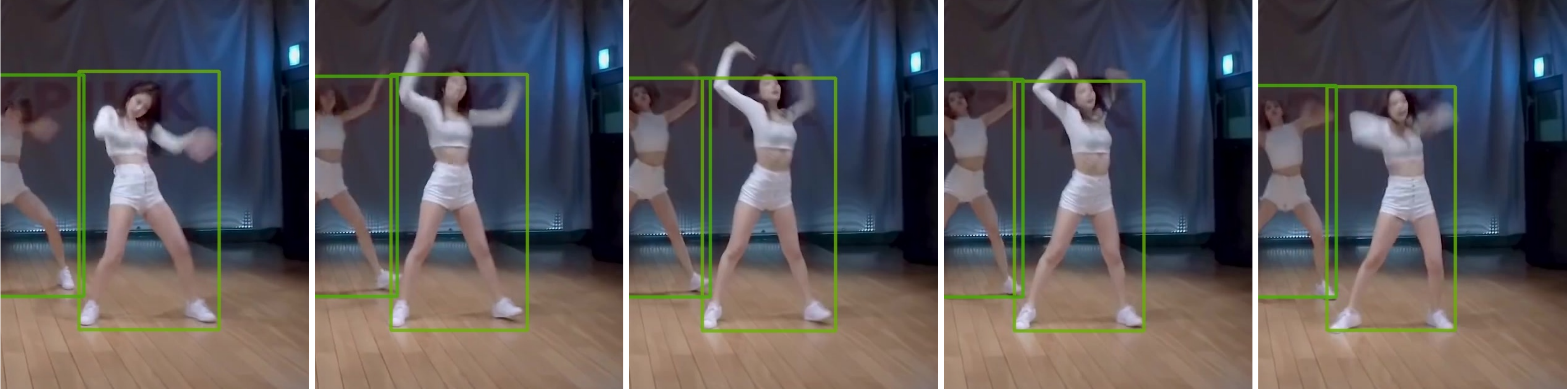}
     \caption{$IoU_{mean}=0.85$, obtained with compression rate 11.7x}
     \label{fig:demo_iou_3}
 \end{subfigure}

   \caption{Visualization of results using different error thresholds, that result in different errors (mean IOU).}
\label{fig:demo_iou}
\end{figure*}

In figure \ref{fig:demo_iou}, we visualize results with different levels of IoU with respect to the ground truth data. As discussed in section 6.1 of the paper, IoU scores lower than 90\%, are visually noticeable. The results on this figure are generated from real tracks in the Dancetrack dataset. We follow the same protocol described in the paper: trajectory simplification, correction of the selected keyframes and interpolation. We use the same hyperparameters, and run our method with different error tolerances $\epsilon_{th}$ to get specific values of $IoU_{mean}$ for the whole dataset.



\section{Qualitative comparison to Uniform sampling}

\begin{figure*}[]
\centering
\begin{subfigure}[b]{0.99\textwidth}
     \centering
     \includegraphics[width=\textwidth,trim={0 0 0 0},clip]{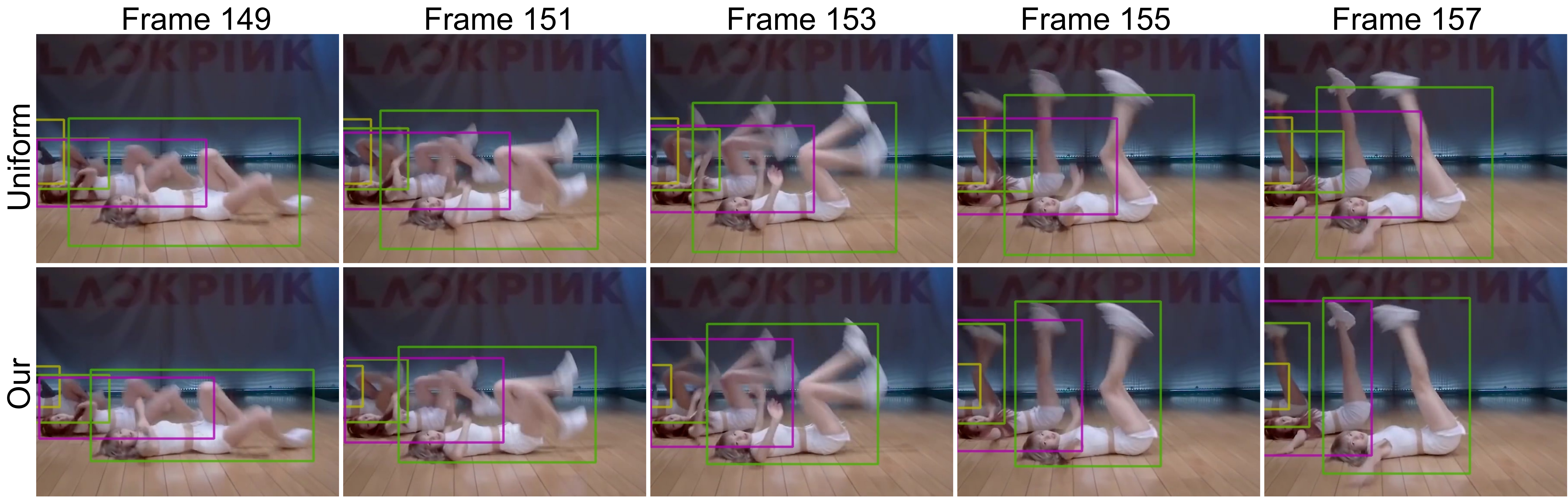}
     \caption{DanceTrack, compression rate 20x}
     \label{fig:demo_video_1}
 \end{subfigure}
 \begin{subfigure}[b]{0.99\textwidth}
     \centering
     \includegraphics[width=\textwidth,trim={0 0 0 0},clip]{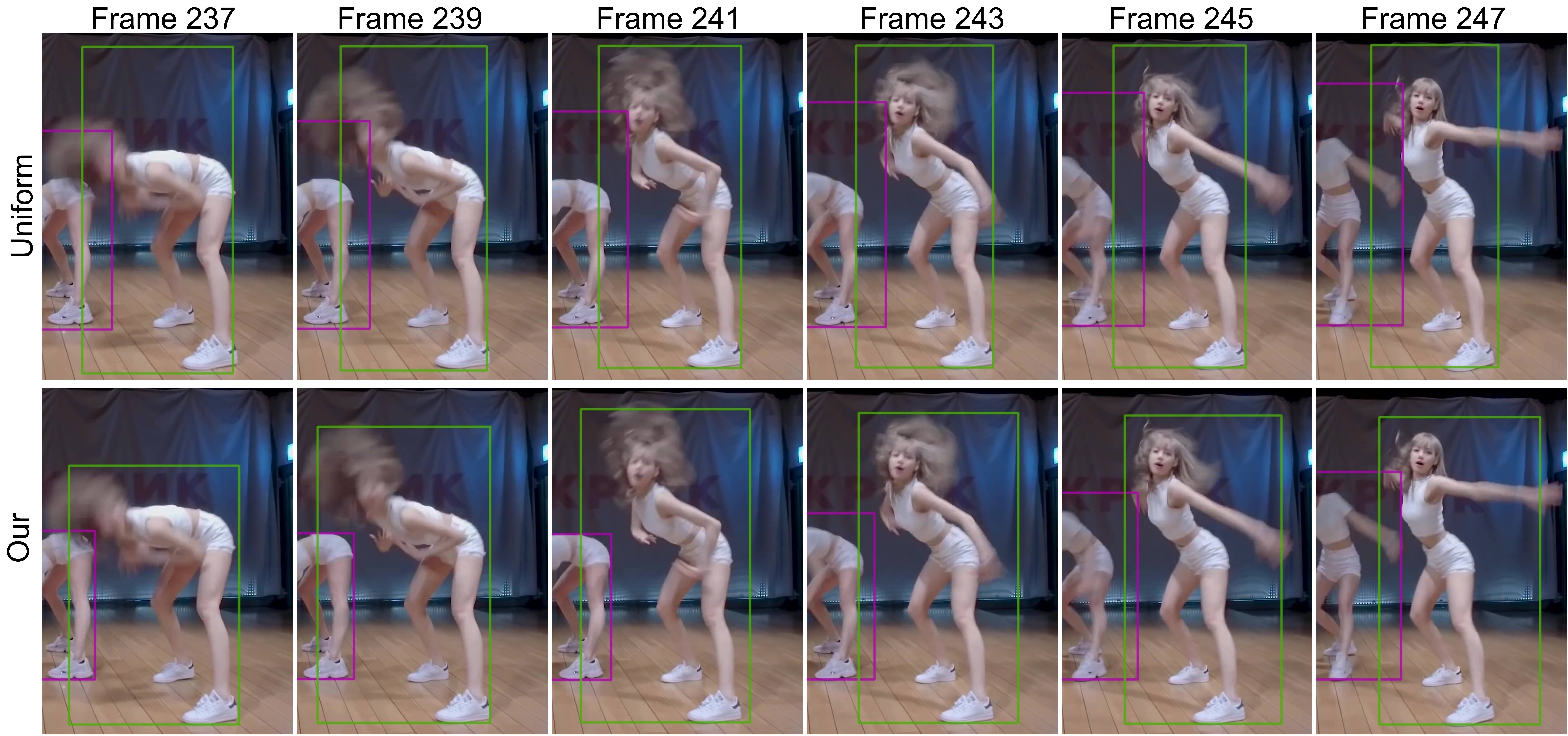}
     \caption{DanceTrack, compression rate 20x}
     \label{fig:demo_video_2}
 \end{subfigure}
 \begin{subfigure}[b]{0.98\textwidth}
     \centering
     \includegraphics[width=\textwidth,trim={0 0 0 0},clip]{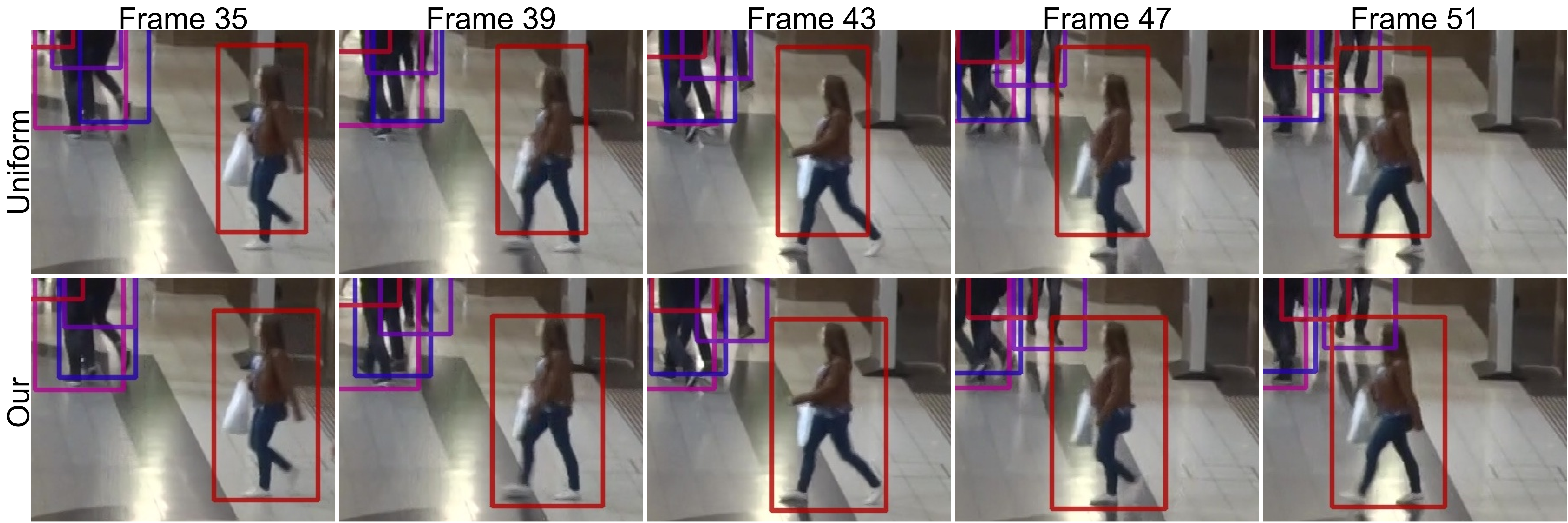}
     \caption{MOT20, compression rate 50x}
     \label{fig:demo_video_3}
 \end{subfigure}

   \caption{Comparison of results using uniform sampling and the proposed method, using the MOT20 and DanceTrack datasets.}
\label{fig:demo_video}
\end{figure*}

Figure \ref{fig:demo_video} compares  uniform sampling and the proposed method at different compression rates, on the DanceTrack and Mot20 datasets. For the DanceTrack dataset, we observe several artifacts with the uniform sampling, specially at 20x compression rate, while the results with the proposed method are visually better. For MOT20, even though the pedestrian motion is mostly linear, small camera movements degrade the performance of uniform sampling at high compression rates, which we do not observe in the proposed method. For a video of the results, refer to the file ``demo\_video\_compressed.mp4" in the supplementary material.    


